\documentclass{article} 
\usepackage[table]{xcolor}
\usepackage{colm2024_conference}
\usepackage{microtype}
\usepackage{hyperref}
\usepackage{url}
\usepackage{booktabs}
\usepackage{amsmath}
\usepackage{multirow}
\usepackage{multicol}
\usepackage{CJKutf8}
\usepackage{tcolorbox}

\definecolor{darkblue}{rgb}{0, 0, 0.5}
\hypersetup{colorlinks=true, citecolor=darkblue, linkcolor=darkblue, urlcolor=darkblue}
\newcommand{\ignore}[1]{}

\title{FinMCP-Bench: Benchmarking LLM Agents for Real-World Financial Tool Use under the Model Context Protocol}

\author{
\begin{tabular}{c}
Jie Zhu\textsuperscript{\textrm{1}}, 
Yimin Tian\textsuperscript{\textrm{1}},
Boyang Li\textsuperscript{\textrm{1}},
Kehao Wu\textsuperscript{\textrm{2}},
Zhongzhi Liang\textsuperscript{\textrm{2}},
Junhui Li\textsuperscript{\textrm{3}}, \\
Xianyin Zhang\textsuperscript{\textrm{1}},
Lifan Guo\textsuperscript{\textrm{1}}\thanks{Corresponding Author.},
Feng Chen\textsuperscript{\textrm{1}},
Yong Liu\textsuperscript{\textrm{2}},
Chi Zhang\textsuperscript{\textrm{1}}
\end{tabular}
\\
\\
\textsuperscript{\textrm{1}}Qwen DianJin Team, Alibaba Cloud Computing \\
\textsuperscript{\textrm{2}}YINGMI Wealth Management \\
\textsuperscript{\textrm{3}}Soochow University
}

\colmfinalcopy 
\begin{document}

\maketitle

\renewcommand{\thefootnote}{}
\footnotetext{This work has been submitted to the IEEE for possible publication. Copyright may be transferred without notice, after which this version may no longer be accessible.}
\renewcommand{\thefootnote}{\arabic{footnote}}

\begin{abstract}
This paper introduces \textbf{FinMCP-Bench}, a novel benchmark for evaluating large language models (LLMs) in solving real-world financial problems through tool invocation of financial model context protocols. FinMCP-Bench contains 613 samples spanning 10 main scenarios and 33 sub-scenarios, featuring both real and synthetic user queries to ensure diversity and authenticity. It incorporates 65 real financial MCPs and three types of samples, single tool, multi-tool, and multi-turn, allowing evaluation of models across different levels of task complexity. Using this benchmark, we systematically assess a range of mainstream LLMs and propose metrics that explicitly measure tool invocation accuracy and reasoning capabilities. FinMCP-Bench provides a standardized, practical, and challenging testbed for advancing research on financial LLM agents.
\end{abstract}

\section{Introduction}
Large language models (LLMs) are increasingly being deployed as agents in financial applications, where they are expected to interpret user requests, call external tools, and carry out multi-step reasoning. In practice, LLM agents must understand user intentions, access financial tools to retrieve information such as stock trends, fund holdings, and market analyses, and then apply financial concepts to generate useful responses. This often requires chaining multiple tool calls together, with each step depending on the results of the previous one. Such implicit dependencies make it difficult to evaluate how well LLM agents handle realistic financial tasks.

While recent work has explored the evaluation of LLMs on general tool use, existing evaluations in the financial domain remain limited to specific tasks and typically do not involve tool use~\cite{lei-etal-2024-cfbenchmark, zhu-etal-2024-cflue, nie-etal-2025-cfinbench, tang-etal-2025-financereasoning, xie-etal-2025-finchain, li-etal-2024-investorbench, m3finmeeting}. To address this gap, we present \textbf{FinMCP-Bench}, a benchmark designed to evaluate LLMs in realistic and challenging financial scenarios through interactions with real-world Model Context Protocol (MCP)~\cite{mcp}, which offer a standardized schema for tool invocation across diverse servers.

Our dataset construction begins with 10K interaction records collected from production financial agents developed by domain experts and deployed in 33 real-world scenarios. These interactions involve 65 financial tools integrated through MCP, covering a wide range of genuine user needs. Each record contains on average more than two tool calls with inherent dependencies. To further increase complexity, we synthesize high-difficulty cases with tool call chains exceeding five steps by leveraging LLM-based augmentation strategies. After expert annotation, we curate a final set of 10K high-quality interaction traces with long tool call chains and strong internal dependencies.

Our main contributions are described as follows:
\begin{itemize}
    \item We propose FinMCP-Bench, a comprehensive benchmark for evaluating LLMs’ ability to invoke MCP tools in financial scenarios. It contains 613 samples across 10 main scenarios and 33 sub-scenarios, including real and synthetic user queries, with three sample types: single tool, multi-tool, and multi-turn, allowing evaluation across different levels of task complexity.

    \item We systematically evaluate a range of mainstream LLMs on FinMCP-Bench and introduce explicit metrics for tool invocation accuracy. Results highlight both the strengths of current models and the challenges they face, particularly in handling complex multi-tool dependencies and multi-turn conversations.
\end{itemize}

\begin{figure}
\centering
\includegraphics[width=1\linewidth]{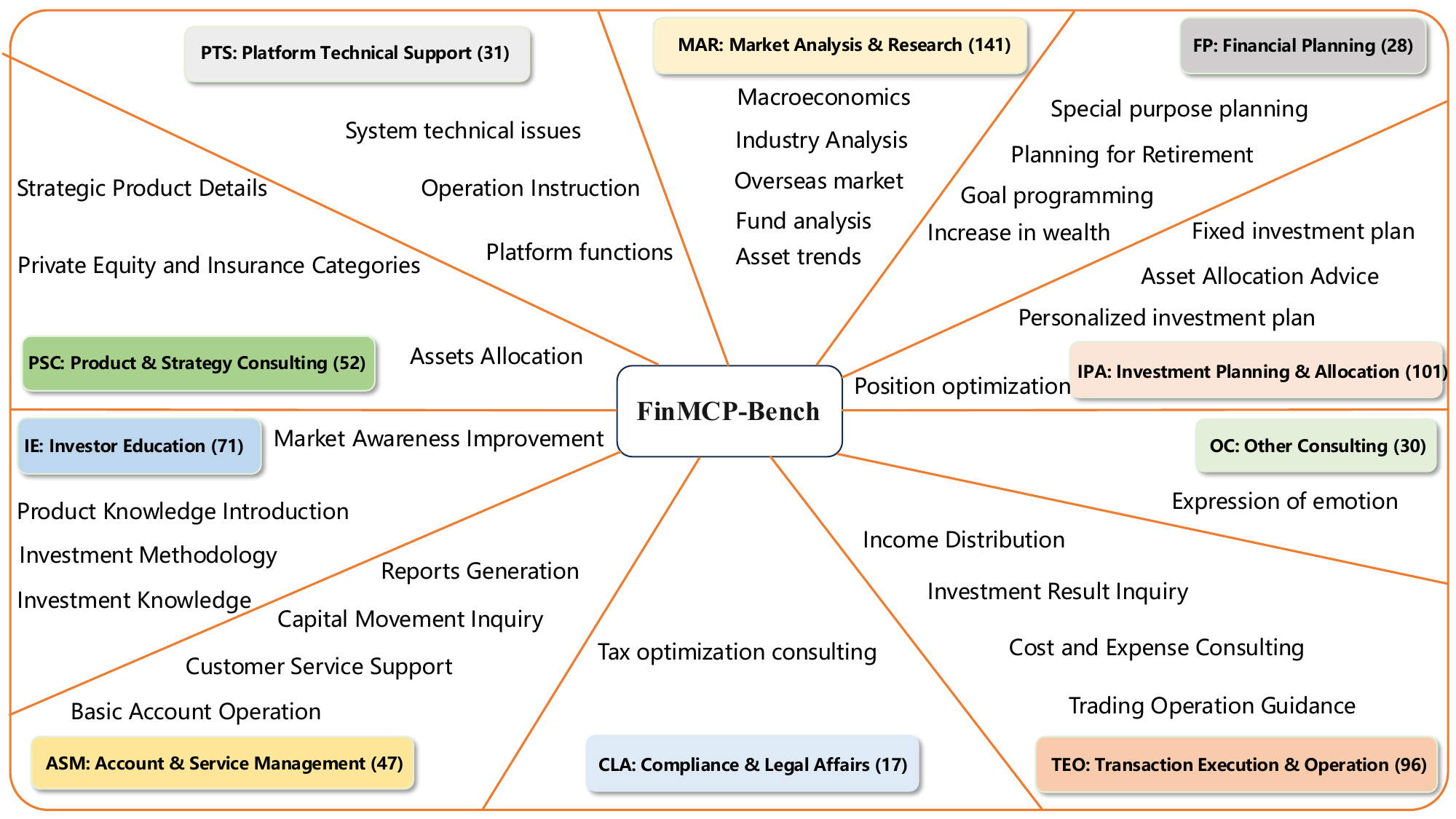}
\caption{The 10 main scenarios and 33 sub-scenarios in FinMCP-Bench.}
\label{fig:category}
\end{figure}

\section{FinMCP-Bench: A Financial MCP Benchmark}

As illustrated in Figure~\ref{fig:category}, FinMCP-Bench contains 613 samples covering 10 main scenarios and 33 sub-scenarios. The categories with the largest number of samples are \textit{Market Analysis and Research (MAR)} with 141 samples and \textit{Investment Planning and Allocation (IPA)} with 101 samples.

We categorize samples based on the complexity and structure of tool usage. Each sample begins with a customer query, which may be addressed in one of three ways:
\begin{itemize}
\item {\bf Single-tool}: resolved with a single tool call in one conversational turn (145 samples).
\item {\bf Multi-tool}: involves multiple tool calls within a single conversational turn, which may be sequential or parallel (249 samples).
\item {\bf Multi-turn}: spans multiple conversational turns, each potentially invoking one or more tools (219 samples).
\end{itemize}

\subsection{Data Source}
We collect high-quality data from 10,000 historical logs of the XiaoGu AI assistant in the Qieman APP operated by Yingmi Fund\footnote{Yingmi Fund is a CSRC-approved fund management and sales company (https://qieman.com).}, where the assistant follows expert-defined SOPs and invokes tools via workflow-style processes. All logs are processed through a strict anonymization and disclosure procedure. To ensure quality, logs are retained only if (i) the query reflects genuine financial needs, (ii) the problem is resolved through tool calls, and (iii) the final response provides a satisfactory solution. From this process, we obtain 1,484 single-tool samples and 183 multi-tool samples. The single-tool samples are then randomly divided into two subsets of 700 and 784. The first subset ($\mathcal{S}$) is reviewed by experts, and 145 high-quality samples are ultimately retained in FinMCP-Bench. The second subset ($\mathcal{S}_{o}$) is reserved for synthesizing more complex multi-tool samples (Section~\ref{sec:multi_step}) and multi-turn samples (Section~\ref{sec:multi_turn}). 

\begin{figure}
    \centering
    \includegraphics[width=1\linewidth]{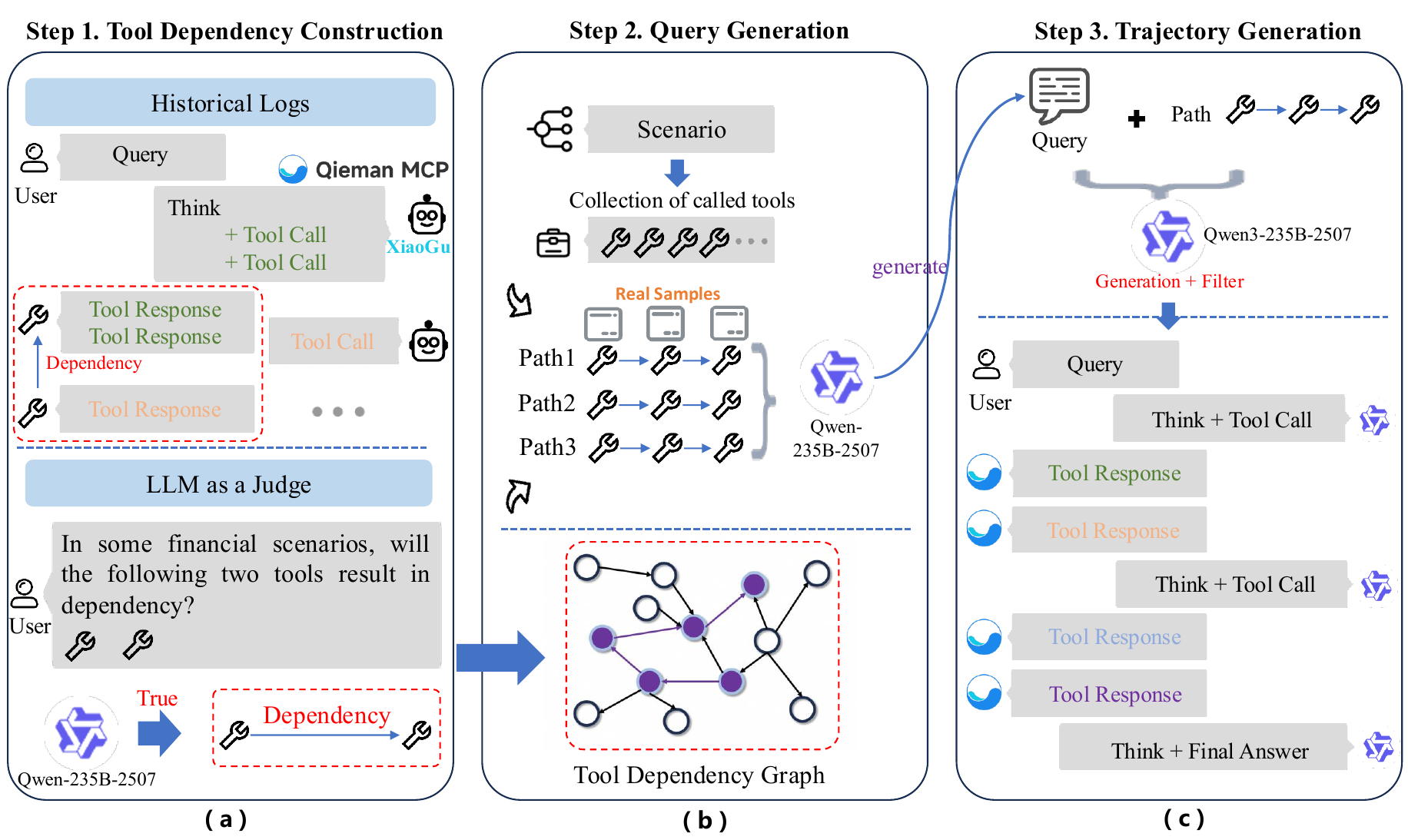}
    \caption{Overview of the chain-based method for synthesizing multi-tool samples.}
    \label{fig:multi_step}
\end{figure}

\subsection{Chain-based Multi-tool Sample Construction}
\label{sec:multi_step}
Figure~\ref{fig:multi_step} illustrates our chain-based method for constructing multi-tool samples, which consists of three stages: (1) building a tool dependency graph, (2) generating user query, and (3) expanding the sampled chains into full trajectories.

\noindent{\bf Tool Dependency Graph Construction.} In this step, we construct the tool dependency graph from scratch by processing each sample in $\mathcal{M}$ through two sub-steps. Initially, the graph contains $N$ tool nodes, $\mathcal{T}=\{t_1,\cdots,t_N\}$, with no edges, where $\mathcal{T}$ denotes the complete tool set. To illustrate the process, consider a multi-tool sample $M \in \mathcal{M}$, where the tools in $M$ are organized into groups. 

In the first sub-step, we obtain pairs of tools that potentially have dependency relation in $M$. If a tool $t_j$ appears in a group that immediately follows a group containing tool $t_i$, we propose a dependency $t_i \rightarrow t_j$. As illustrated in the top of Figure~\ref{fig:multi_step}(a), three tools are arranged into two groups: $(t_1, t_2)$ and $(t_3)$. This yields two candidate dependencies, $t_1 \rightarrow t_3$ and $t_2 \rightarrow t_3$. Since $t_1$ and $t_2$ are invoked in parallel, no dependency is assumed between them. In the second sub-step, shown at the bottom of Figure~\ref{fig:multi_step}(a), we verify the validity of each candidate pair $t_i \rightarrow t_j$ using an LLM (Qwen3-235B-2507). The model is prompted to judge whether it is reasonable for $t_j$ to depend on $t_i$. If validated, we add a directed edge from $t_i$ to $t_j$ in the dependency graph. 

The final tool dependency graph $\mathcal{G}$, as shown in the bottom of Figure~\ref{fig:multi_step}(b) contains 65 nodes with 288 edges.

\noindent{\bf User Query Generation.} 
For each scenario, we begin by sampling tool pairs $(t_i, t_j)$ from the tool set $\mathcal{T}$, ensuring that both tools appear in samples from $\mathcal{M}$ associated with the same scenario. If multiple paths exist between $t_i$ and $t_j$ in the tool dependency graph $\mathcal{G}$, we randomly select one, which defines a tool chain from $t_i$ to $t_j$.

Given a tool chain $\mathcal{C}={c_1, \cdots, c_n \mid c_i\in\mathcal{T}}$ containing $n$ tools, we generate a user query that aligns with this chain. For each tool $c_i$ in $\mathcal{C}$, we randomly select a single-tool sample $s_i$ from $\mathcal{S}_o$ such that $s_i$ involves $c_i$. Using the set $\{s_1, \cdots, s_n\}$ as in-context examples, we prompt Qwen3-235B-2507 to generate a proper user query.

\noindent{\bf Trajectory Generation.} As shown in the top of Figure~\ref{fig:multi_step}(c), Qwen3-235B-2507, connected to Qieman’s MCP server, is then used to generate the corresponding trajectory for each user query. A query-trajectory pair is retained as a multi-tool sample if its trajectory (i) may include additional tools beyond those in $\mathcal{C}$, and (ii) correctly preserves the dependency relations specified in $\mathcal{C}$.

In practice, we generate 1K query-trajectory pairs, from which 496 multi-tool samples are obtained. These, combined with the 183 multi-tool samples extracted from real logs, are then manually reviewed by experts. After this review process, 249 high-quality multi-tool samples are ultimately retained in FinMCP-Bench.

\begin{figure}
    \centering
    \includegraphics[width=1.0\linewidth]{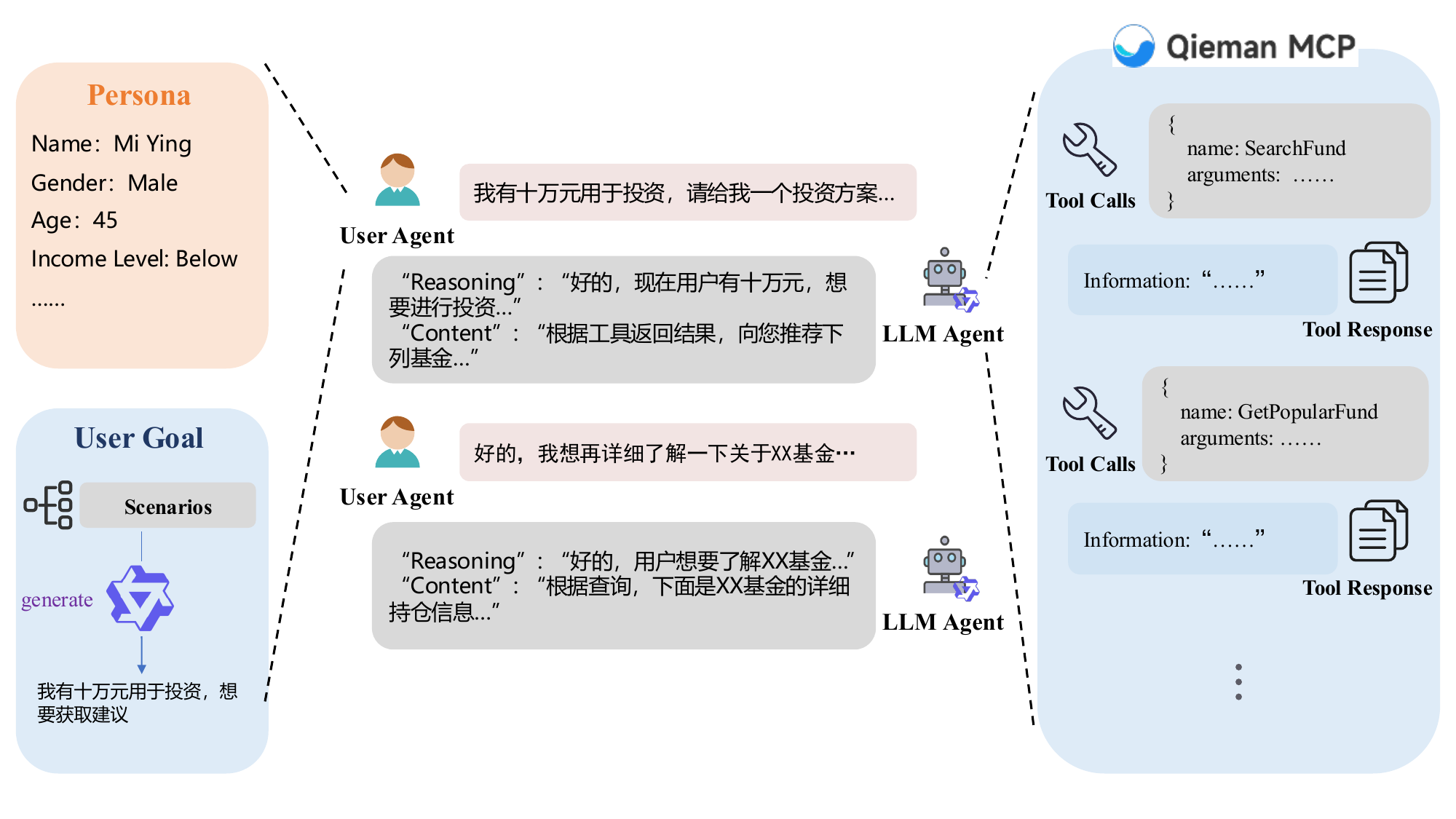}
    \caption{Overview of the role-playing-based method for synthesizing multi-turn samples.}
    \label{fig:multi_turn}
\end{figure}

\subsection{Role-Playing-based Multi-turn Sample Construction}
\label{sec:multi_turn}

Next, we construct multi-turn samples in which tools are invoked over several rounds of dialogue between a user and a service assistant. To mimic realistic conversations, we design a dialogue framework where a planner agent specifies both the user persona and the user goal, as illustrated on the left side of Figure~\ref{fig:multi_turn}.
\begin{itemize}
\item {\bf Persona}. We sample user personas from the character profile pool introduced by~\cite{zhu-etal-2025-evaluating}, which provides a comprehensive template for financial customers. The template includes attributes such as age, gender, and income level, all of which are highly relevant in real-world financial contexts. 
\item {\bf User Goal}. The planner first selects a sub-scenario and, together with the chosen persona, prompts Qwen3-235B-2507 to generate a corresponding user goal. 
\end{itemize}

Once the user persona and task instruction are defined, we simulate the dialogue by assigning Qwen3-235B-2507 to play both the {\it user} and the {\it assistant} roles, as illustrated on the right side of Figure~\ref{fig:multi_turn}. 

We generate 500 multi-turn dialogues as candidate samples. To ensure quality, we first use Qwen3-235B-2507 to automatically check their validity, with a focus on whether all user queries in each dialogue are successfully addressed. This filtering step reduces the set to 378 dialogues, which are then manually reviewed by financial experts. After this expert review, 219 high-quality multi-turn samples are retained in FinMCP-Bench.

\subsection{Quality Control}
To ensure the quality of the benchmark, we invite six domain experts and experienced developers in the financial field to evaluate both real and synthetic samples. Specifically, we use a two-stage pipeline: automated validation and expert review. In the first stage, an automated validator checks whether all tools are executed successfully without errors. In the second stage, six financial experts serve as reviewers. Each sample is independently evaluated by two randomly assigned experts, who score it on a 5-point Likert scale~\cite{joshi-etal-2015-likert} across five dimensions: question relevance, tool-chain completeness, tool-chain logical consistency, answer reliability and traceability, and data freshness. A sample is accepted only if both reviewers assign a score of at least four in all dimensions. When the two reviewers disagree, the sample is resolved through discussion.

\begin{table}[t!]
\footnotesize
\centering
\begin{tabular}{ll}
\hline
\textbf{Type} & \textbf{Number} \\
\hline
\multicolumn{2}{c}{\textit{FinMCP-Bench dataset}} \\
All  & 613\\
\quad \textit{- Easy/Medium/Hard} & 343 / 245 / 25\\
Single-Tool  & 145 \\
Multi-Tool  & 249 \\
\quad \textit{- MCP calls per instance (Median / Avg)} & 8 / 7.32 \\
\quad \textit{- Steps per instance (Median / Avg)} & 5 / 5.72 \\
\quad \textit{- \#Samples having Parallel Calls} & 73 \\

Multi-Turn & 219 \\
\quad \textit{- MCP calls per instance (Median / Avg)} & 4 / 5.00 \\
\quad \textit{- Number of Turns (Median / Avg)} & 6 / 5.95 \\
\hline
\multicolumn{2}{c}{\textit{Qieman MCP}} \\

Tool Number & 65 \\
Arguments \textit{(Avg)} & 2.6\\
\hline
\end{tabular}
\caption{Statistics of FinMCP-Bench.}
\label{tab:statistics}
\end{table}

\subsection{Dataset Analysis}
Without loss of generality, the tools invoked in a sample can be represented as
$\{(t^1_1, \dots, t^1_{n_1}), \dots, (t^M_1, \dots, t^M_{n_M})\}$,
where there are $M$ groups of tools and the $i$-th group contains $n_i$ tools. Tools within the same group are executed in parallel, so their order is interchangeable. In particular, single-tool samples correspond to the special cases where $M=1$ and $n_1=1$. As shown in Table~\ref{tab:statistics}, FinMCP-Bench contains 613 samples that differ in difficulty based on the number of tool calls required. For simplicity, we categorize samples with up to 5 tool calls as {\it easy}, those with up to 10 tool calls as {\it medium}, and the remaining samples as {\it hard}.

All 145 single-tool samples contain exactly one tool call in a single step. In contrast, multi-tool samples contain on average 7.32 tool calls across 5.72 steps. Among these, 73 out of 249 multi-tool samples include parallel calls, where multiple tools are invoked within a single step. Multi-turn samples, on average, contain span 5.95 conversational turns and invoke 5.00 tools.

\section{Experimentation}

\begin{table*}[t!]
\centering
\resizebox{\linewidth}{!}{
\begin{tabular}{l l l l l l l l l l l l l l l l l}
\toprule
\textbf{Model} & \multicolumn{4}{c}{\textbf{Single-Tool}} & \multicolumn{4}{c}{\textbf{Multi-Tool}} & \multicolumn{4}{c}{\textbf{Multi-Turn}} & \multicolumn{4}{c}{\textbf{All}} \\
\cmidrule(lr){2-5} \cmidrule(lr){6-9} \cmidrule(lr){10-13} \cmidrule(lr){14-17} & \textbf{TP} & \textbf{TR} & \textbf{TF1} & \textbf{EMR} & \textbf{TP} & \textbf{TR} & \textbf{TF1} & \textbf{EMR} & \textbf{TP} & \textbf{TR} & \textbf{TF1} & \textbf{EMR} & \textbf{TP} & \textbf{TR} & \textbf{TF1} & \textbf{EMR}  \\
\midrule
DeepSeek-R1 & \underline{59.87} & 65.28 & 62.46 & 49.31 & 65.95 & 43.42 & 52.36 & 4.85 & 46.15 & 5.15 & 9.27 & 0.00 & \bf 64.79 & 40.55 & 49.88 & 18.08 \\
GPT-OSS-20B & 13.50 & 68.97 & 22.57 & 15.17 & 28.26 & \underline{60.59} & 38.54 & 1.20 & 9.51 & 6.03 & 7.38 & 0.00 & 25.97 & 43.84 & 32.62 & 4.43  \\
Seed-OSS-36B & 35.59 & 72.41 & 47.73 & 35.17 & 58.21 & 28.13 & 37.93 & 3.61 & 39.26 & 45.49 & 42.15 & 2.17 & 50.36 & 32.28 & 39.34 & 13.86 \\
Qwen3-4B-Thinking & \bf 63.01 & \underline{75.17} & \bf 68.55 & \bf 65.52 & 64.44 & 41.16 & 50.23 & 4.82 & \underline{49.23} & \underline{46.16} & \bf 47.65 & 2.26 & 58.02 & 44.06 & 50.08 & \underline{18.82} \\
Qwen3-30B-A3B-Thinking & 40.96 & 70.34 & 51.78 & 55.17 & \bf 67.92 & 54.92 & \underline{60.73} & \underline{7.47} & \bf 53.17 & 41.17 & \underline{46.40} & \bf 4.10 & \underline{61.75} & \underline{50.53} & \underline{55.58} & 18.24 \\
Qwen3-235B-A22B-Thinking & 55.76 & \bf 83.45 & \underline{66.85} & \underline{60.00} & \underline{67.02} & \bf 72.00 & \bf 69.42 & \bf 10.62 & 36.50 & \bf 48.25 & 41.56 & \underline{3.08} & 60.22 & \bf 68.90 & \bf 64.27 & \textbf{25.92} \\
\bottomrule
\end{tabular}
}
\caption{Results (\%) on FinMCP-Bench.}
\label{tab:result}
\end{table*}

\subsection{Experimental Settings}
\noindent{\bf Models.} We evaluate six large language models, including three from the Qwen3 family ~\cite{qwen3}: Qwen3-4B-Thinking, Qwen3-30B-A3B-Thinking, and Qwen3-235B-A22B-Thinking, as well as three additional models: DeepSeek-R1 ~\cite{deepseekr1}, GPT-OSS-20B ~\cite{gpt-oss}, and Seed-OSS-36B ~\cite{seed-oss}.

\noindent{\bf Inference.} Single-tool and multi-tool samples can be treated as one-turn conversations, consisting of a user query and an agent reply, while multi-turn samples naturally represent multi-turn conversations. Without loss of generality, we denote a conversation as $\{(u_1, r_1), \dots, (u_n, r_n)\}$, where $n$ is the number of turns and each reply $r_i$ includes both tool calls and responses. Following the task customer support conversation~\cite{zhu-etal-2025-evaluating}, we treat the LLM as the agent. For each turn $i$, the model is prompted to generate a reply $r_i^{\prime}$ given the current user utterance $u_i$ and the gold conversation history $\{(u_1, r_1), \dots, (u_{i-1}, r_{i-1})\}$. From the generated replies $\{r_1^{\prime}, \dots, r_n^{\prime}\}$, we can extract the tools invoked by the model.

\subsection{Evaluation Metrics} 
Unlike previous work that focuses on the accuracy of the final answer ~\cite{dong2025toolstarempoweringllmbrainedmultitool, li2025torlscalingtoolintegratedrl, qian2025toolrlrewardtoollearning,goldie2025syntheticdatageneration}, in this paper we evaluate LLM performance based on the tools invoked and propose the following metrics:\footnote{In financial research and advisory scenarios, queries are inherently open-ended without standard answers; hence, evaluation focuses on tool-use capability.}

\noindent{\bf Tool Recall (TR).} We construct a reference tool set and a predicted tool set by extracting tools from the reference and the prediction, while ignoring dependency relations. {\bf Tool Recall} is defined as the number of correctly predicted tools (i.e., those appearing in both sets) divided by the total number of tools in the reference set.

\noindent{\bf Tool Precision (TP).} Similarly, we define {\bf Tool Precision}, as the number of correctly predicted tools (i.e., those appearing in both sets) divided by the total number of tools in the predicted set.

\noindent{\bf Tool F1 (TF1).} To balance Tool Precision (TP) and Tool Recall (TR), we define {\bf Tool F1} as their harmonic mean: $\text{TF}_1=\frac{2\times \text{TR} \times \text{TP}}{\text{TR} + \text{TP}}$.

\noindent{\bf Exact Match Rate (EMR).} Unlike the previous metrics, which compare tools without considering their grouping, the strictest way to evaluate prediction accuracy is to check whether the predicted tool organization exactly matches the reference. Since tools within the same group can be invoked in parallel, their internal order is ignored. The proportion of predictions that exactly match the reference organization is defined as the {\bf Exact Match Rate (EMR)}.

\begin{figure}[t!]
    \centering
    \begin{minipage}[t]{0.48\linewidth}
        \centering
        \includegraphics[width=\linewidth]{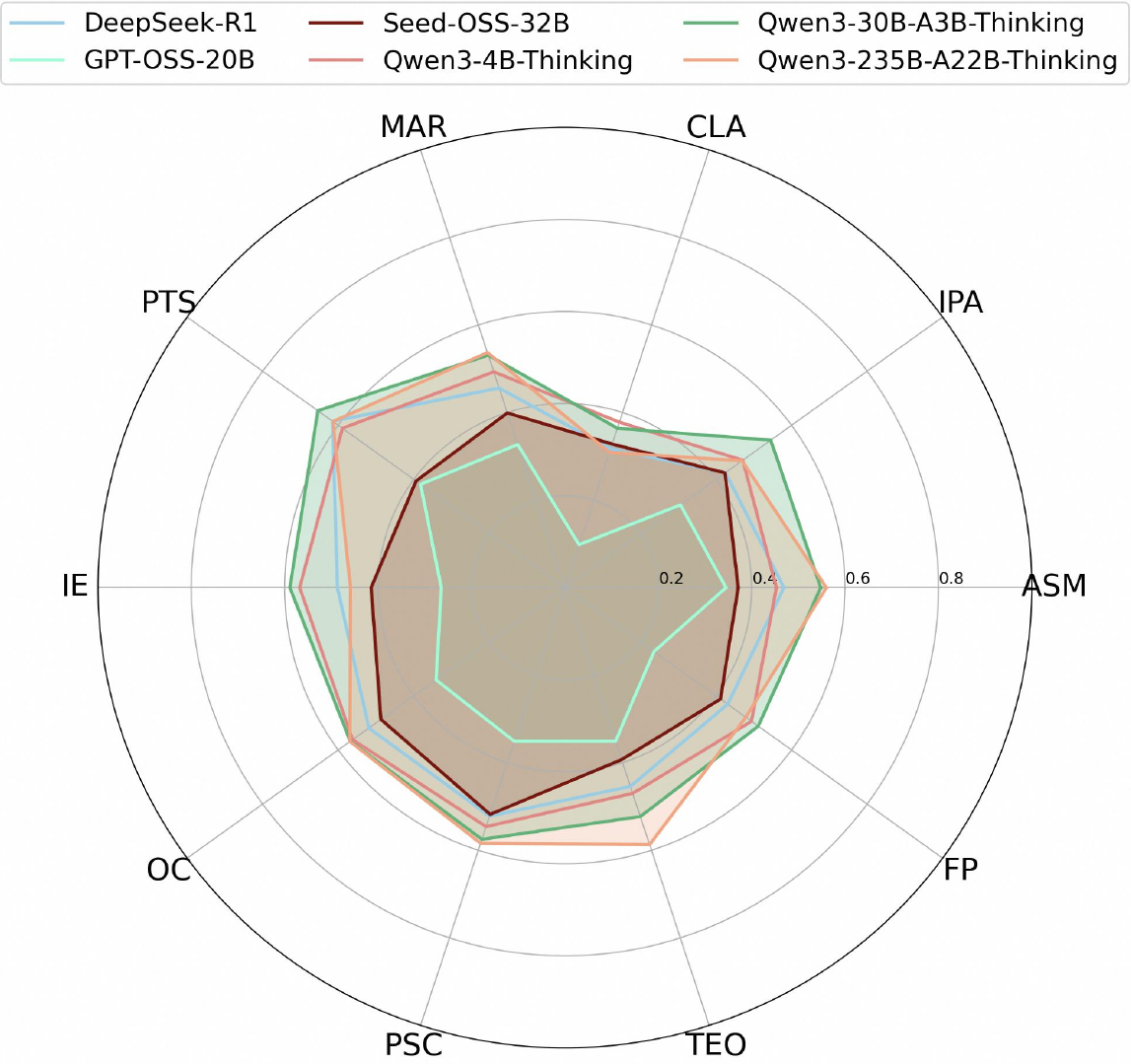}
        \caption{Scenario-wise TF1 results on FinMCP-Bench. Full scenario names are listed in Figure~\ref{fig:category}.}
        \label{fig:scenario_result}
    \end{minipage}
    \hfill
    \begin{minipage}[t]{0.48\linewidth}
        \centering
        \includegraphics[width=\linewidth]{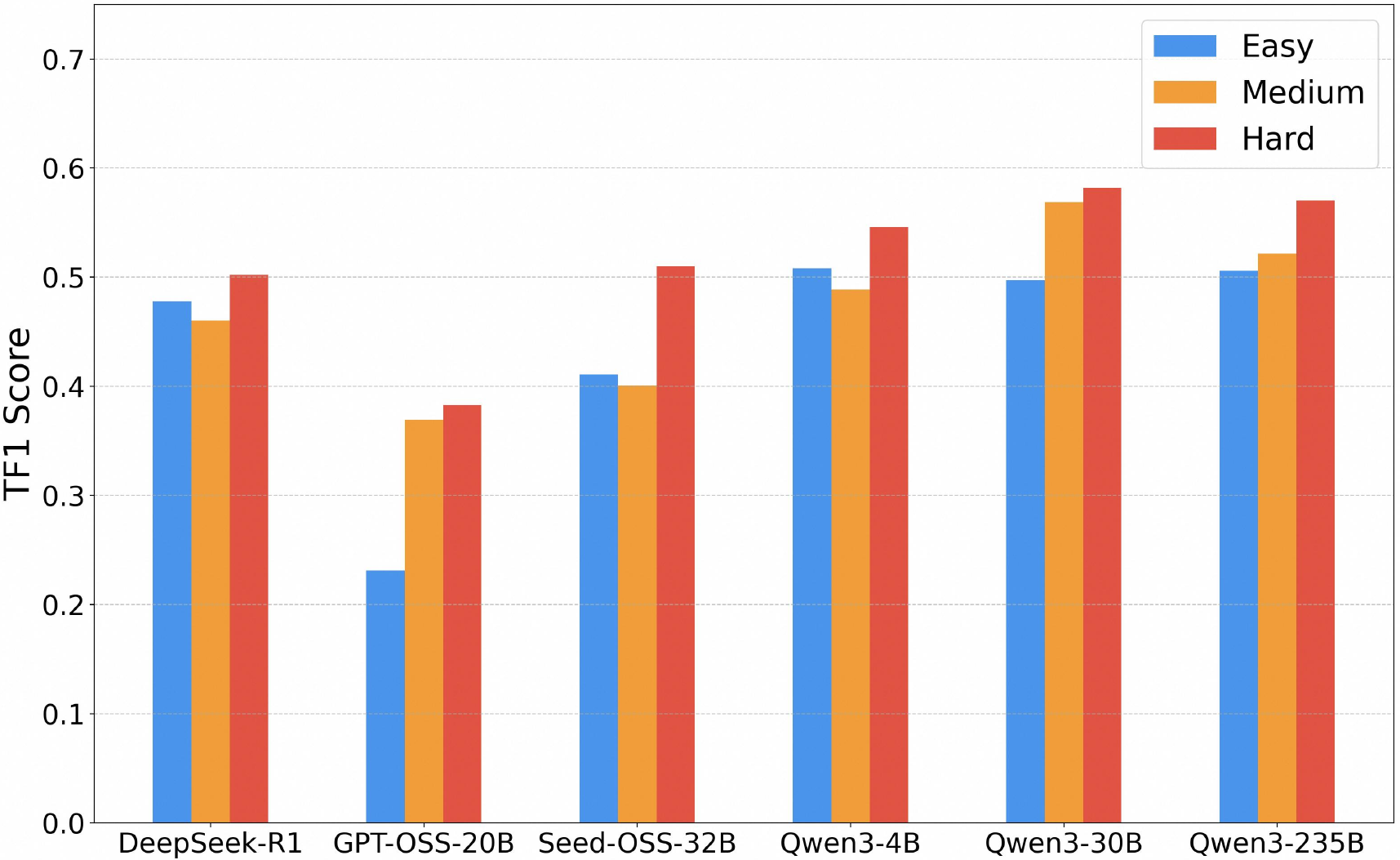}
        \caption{Difficulty-wise TF1 on FinMCP-Bench}
        \label{fig:diff_result}
    \end{minipage}
\end{figure}

\subsection{Experimental Results}
Table~\ref{tab:result} reports model performance across single-tool, multi-tool, and multi-turn samples. We make the following observations:
\begin{itemize}
\item Overall, the three Qwen3 models generally outperform the others on both TF1 and EMR. However, model size does not consistently correlate with performance: Qwen3-4B-Thinking achieves a higher EMR score than Qwen3-30B-A3B-Thinking, while Qwen3-30B-A3B-Thinking attains a higher TF1 score than Qwen3-4B-Thinking.

\item Comparing single-tool and multi-tool samples, we find that Tool Recall (TR) is higher for single-tool samples, as each contains only one tool. In contrast, Tool Precision (TP) is lower for single-tool samples, since models often over-predict by generating multiple tools even when only one is needed.

\item Multi-turn samples tend to yield the lowest scores overall, especially in EMR, indicating that handling longer conversations with multiple tool calls remains challenging. 
\end{itemize}

\subsection{Experimental Analysis}
\noindent{\bf Scenario-wise Results} The radar chart of Figure~\ref{fig:scenario_result} shows ther TF1 performance with respective to the ten main scenarios. It shows that Qwen3-30B-A3B-Thinking and Qwen3-235B-A22B-Thinking form the leading group with the largest, most rounded profiles, indicating strong and balanced tool use across scenarios. Qwen3-4B-Thinking is a solid second tier, while DeepSeek-R1 and Seed-OSS-36B are mid-range with noticeable dips. GPT-OSS-20B lags across all axes. Performance gaps widen in scenarios requiring multi-tool planning and cross-source synthesis, but narrow on simpler, single-operation queries. Overall, the top models lead by maintaining a better precision-recall balance across diverse scenarios rather than excelling in only a few.

\noindent{\bf Difficulty-wise Results} Across Easy, Medium, and Hard splits, TF1 does not monotonically decline with difficulty. Stronger models (Qwen3-30B-A3B-Thinking and Qwen3-235B-A22B-Thinking) improve from Easy to Hard, suggesting they leverage richer constraints and multi-tool opportunities in harder queries. Qwen3-4B-Thinking shows a mild upward trend, while DeepSeek-R1 and Seed-OSS-36B rise modestly. GPT-OSS-20B exhibits a large jump from Easy to Medium/Hard but remains behind others. Overall, easy cases penalize over-calling (lower precision), whereas harder cases reward better recall and planning, yielding higher TF1 for models with balanced tool selection.

\section{Conclusion}
In this paper, we present FinMCP-Bench, a new benchmark for evaluating LLMs in real-world financial scenarios that require invoking MCP tools. The benchmark covers three categories of tasks, single-tool, multi-tool, and multi-turn, capturing different levels of complexity in tool usage and dialogue interaction. We conduct extensive evaluations of several popular LLMs on FinMCP-Bench and analyze their performance across multiple dimensions. The results highlight both the strengths of current models and the challenges they face, particularly in handling complex multi-tool dependencies and multi-turn conversations. 

We hope FinMCP-Bench can serve as a standardized and challenging testbed for advancing research on tool-augmented LLMs in finance and inspire future work on improving reasoning, tool orchestration, and dialogue capabilities in this critical domain.

\bibliography{colm2024_conference}
\bibliographystyle{colm2024_conference}

\appendix

\end{document}